\documentclass{article} 
\usepackage{iclr2025_conference,times}

\usepackage{amsmath,amsfonts,bm}

\def\eqref#1{equation~\ref{#1}}

\def\1{\bm{1}}

\DeclareMathAlphabet{\mathsfit}{\encodingdefault}{\sfdefault}{m}{sl}
\SetMathAlphabet{\mathsfit}{bold}{\encodingdefault}{\sfdefault}{bx}{n}

\usepackage{hyperref}
\usepackage{url}
\usepackage{booktabs}
\usepackage{tabularx}
\usepackage{pifont}

\title{Sycophancy Claims about Language Models: \\ The Missing Human-in-the-Loop}
\author{
Jan Batzner$^{1,3}$, Volker Stocker$^{1,2}$, 
Stefan Schmid$^{1,2}$, 
Gjergji Kasneci$^{3}$
\\
$^{1}$Weizenbaum Institute 
$^{2}$Technical University Berlin \\
$^{3}$Munich Center for Machine Learning \& Technical University Munich 
\\
\texttt{jan.batzner@weizenbaum-institut.de}
}

\iclrfinalcopy 

\begin{document}

\maketitle

\begin{abstract}
Sycophantic response patterns in Large Language Models (LLMs) have been increasingly claimed in the literature. We review methodological challenges in measuring LLM sycophancy and identify five core operationalizations. Despite sycophancy being inherently human-centric, current research does not evaluate human perception. Our analysis highlights the difficulties in distinguishing sycophantic responses from related concepts in AI alignment and offers actionable recommendations for future research.
\end{abstract}

\section{Etymology: From Trading Figs to AI Model Evaluation}
Sycophancy describes an undesired form of flattery or fawning in a servile or insincere way, especially to gain favor \citep{lofberg1917sycophancy}. While the term has gained prominence in contemporary AI research, its origins trace back to ancient Greece. The word derives from the Greek `sukophantēs', combining `sykos' (fig) and `phainein' (to show or reveal), as detailed by \citep[p. 426]{damico2018law}. This etymology reflects its origins in Athenian commerce law, specifically regulations around fig exports, where the term evolved to describe those who leveraged false accusations for personal advantage \citep{harvey1990sycophant,osborne1990vexatious}. This historical conception of sycophancy as calculated insincerity for personal gain assumes human agency and motivation, making it inherently opportunistic and human-centric.

\section{The Many Faces of Sycophancy in AI Alignment Research}
In AI alignment research, the term \textit{sycophancy} has been used to describe a specific form of undesirable model behavior\footnote{Throughout this paper, we adopt anthropomorphic terminology as it appears in the related work we reference, while acknowledging its inherent limitations.}. More specifically, language models are considered sycophantic if they adapt their output to please users, even when such responses are flawed or incorrect \citep{perez2022discoveringlanguagemodelbehaviors, sharmatowards}. Although motivated by AI safety concerns, such as creating ``echo chambers by repeating users' preferred answers'' \citep{perez2022discoveringlanguagemodelbehaviors}, distinguishing the concept from related AI alignment concepts like \textit{personalization} is unclear \citep{batzner2024germanpartiesqabenchmarkingcommerciallarge}. \citet{perez2022discoveringlanguagemodelbehaviors} introduced sycophancy as a systematic bias resulting from reinforcement learning from human feedback (RLHF), an alignment approach in which models learn to optimize for human approval, but not necessarily truthful or helpful responses \citep{chen2024openworldevaluationretrievingdiverse}. This phenomenon manifests itself in various ways. For example, models may conform to user biases or exhibit increased susceptibility to deceptive prompts \citep{zhao2024analyzingmitigatingsycophancylarge}. In retrieval contexts, systems exhibit sycophancy by preferentially surfacing information that aligns with the perspective in the query \citep{chen2024openworldevaluationretrievingdiverse}. Some researchers have reframed this behavior and conceptualized it as (i) `specification gaming' \citep{denison2024sycophancysubterfugeinvestigatingrewardtampering}, highlighting how AI systems may learn unintended behaviors that are inadvertently rewarded during training, or as (ii) `agreeableness bias' \citep{lim2024measuringagreeablenessbiasmultimodal}. This conceptual ambiguity of sycophancy in AI alignment research is further evidenced by \citet{lim2024measuringagreeablenessbiasmultimodal}'s terminological shift from `sycophancy' to `agreeableness bias' between their August and October 2024 preprint versions. While conceptualizations of sycophancy vary across the AI alignment literature, researchers share a common concern: alignment processes may inadvertently cause models to prioritize user approval at the expense of factual or otherwise more balanced outputs.

\section{Measuring Sycophancy: Claims and Requirements}
Although sycophantic behavior is frequently reported in AI alignment research, the terminology lacks systematic definition and classification. We review studies that quantify sycophancy in language models, presenting their methodological approaches and evaluation frameworks. We identify five main measurement approaches: persona-based prompts (``I am''/``You are''), direct questioning (``Are you sure?''), keyword/query-based manipulation, visual misdirection, and LLM-based evaluations (Table \ref{tab:sycophancy_strategies}). These approaches have been evaluated using various benchmarks, including multiple choice tasks, free-form text evaluation, vision language QA, and retrieval diversity testing (Table \ref{tab:sycophancy_strategies}).

While sycophancy implies behavior intended to gain human approval, current research methods largely evaluate this phenomenon without direct human involvement in the assessment process. The reviewed papers use persona-based and non-persona-based evaluation approaches. The persona-based approach \citep{perez2022discoveringlanguagemodelbehaviors,wei2024simplesyntheticdatareduces,denison2024sycophancysubterfugeinvestigatingrewardtampering} uses role descriptions (e.g., ``I am a 38 year old PhD candidate in computer science at MIT''; \citet[p. 2]{perez2022discoveringlanguagemodelbehaviors}) to evaluate model responses. In contrast, non-persona-based approaches employ techniques such as direct questioning (``Are you sure?''; \citet{sharmatowards,chen2024yesmentruthtellersaddressingsycophancy}) or prompt-based misleading approaches (e.g., \citet{zhao2024analyzingmitigatingsycophancylarge,rrv2024chaoskeywordsexposinglarge,ranaldi2024largelanguagemodelscontradict,lim2024measuringagreeablenessbiasmultimodal}) to assess sycophantic tendencies. Although \citet{williams2024multiobjectivereinforcementlearningai} is the only work in our sample that incorporates human evaluation through crowdworkers, their assessment focused on overall model performance rather than on specifically measuring human perception of sycophantic behavior.

Synthesizing the previous insights reveals a critical methodological gap between the claims made about sycophancy in language models and current evaluation approaches. This disconnect raises fundamental questions about the validity of existing research designs and their different approaches in how they conceptualize sycophancy in the context of AI alignment research. Although automated evaluations allow for scalable assessment frameworks, three limitations emerge:
\textit{First}, they may not be able to comprehensively capture the ways in which language models adapt their responses to seek human approval. This is primarily due to the lack of a direct assessment of human perception. \textit{Second}, they may not be able to disambiguate and precisely infer the factors that shape model behavior and their influence. \textit{Third}, they rely on different conceptualizations of sycophancy, thus inherently limiting cross-study comparability.

\section{Conclusion and Recommendations}
Our analysis reveals a fundamental disconnect in sycophancy research: while the term describes behavior intended to gain human approval, current measurement approaches lack a coherent understanding of `AI sycophancy' as well as a direct assessment of human perception. Despite the proliferation of automated metrics, benchmarks, and evaluation frameworks (Table \ref{tab:sycophancy_strategies}), none of those we reviewed explicitly measures how humans perceive sycophantic language model behavior.
This methodological gap raises critical questions about the validity and comparability of current sycophancy evaluation research designs. Future research should prioritize the following:
\begin{itemize}
\item[\ding{51}] \textbf{Terminology:} Development of a coherent understanding of `AI sycophancy' to enable consistent measurement and cross-study comparability.
\item[\ding{51}] \textbf{Human-Centricity:} To claim sycophancy, develop methodological frameworks for measuring human perceptions, consistent with the human-centric assumptions underlying the concept of sycophancy.
\item[\ding{51}] \textbf{Specificity:} When evaluating model responses without human perception, use terminology like ``agreeableness bias'' or ``response alignment'' that better reflects the concept being measured.
\end{itemize}
Addressing these definitional and methodological challenges is crucial to establish coherent metrics of AI sycophancy and to distinguish it meaningfully from related concepts such as personalization.

\subsection*{Acknowledgements}
Jan Batzner, Volker Stocker, and Stefan Schmid acknowledge funding by the German Federal Ministry of Education and Research (BMBF) under grant no. 16DII131 (Weizenbaum-Institut fuer die vernetzte Gesellschaft – Das Deutsche Internet-Institut). Stefan Schmid acknowledges funding by the German Research Foundation (DFG), project ReNO (SPP 2378), 2023-2027. We thank Monserrat López Pérez, Gabriel Freedman, Andrew Caunes, Kaushik Sanjay Prabhakar, and Jonathan Reti. The authors acknowledge that the first author meets the URM criteria of the ICLR 2025 Tiny Papers Initiative.


\nocite{*}
\bibliography{iclr2025_conference}
\bibliographystyle{iclr2025_conference}
\vspace{5cm}

\appendix
\section*{Comparison of common sycophancy measurement approaches}
\begin{table*}[ht]
\footnotesize
\centering
\small
\begin{tabular}{p{1.9cm}p{3.5cm}p{3cm}p{3.5cm}p{1.8cm}}
\toprule
\textbf{Measurement Approach} & \textbf{Core Mechanism} & \textbf{Opportunties} & \textbf{Challenges} & \textbf{Sample \newline References} \\ 
\midrule
\textbf{Persona Prompts} 
& 
\textit{Inject a synthetic or real-world persona} (e.g., ``I am a 38-year-old PhD candidate''), then observe whether the model adjusts its answers to align with that persona. 
& 
(i) Control for isolated persona attributes;
(ii) Enables counterfactual experiments;  
(iii) Reflect Role Playing via system prompts. 
& 
(i) Representativeness and ecological validity of those personae; 
(ii) May conflate personalization with sycophancy; 
(iii) Selection biases in persona design. 
& 
\citet{perez2022discoveringlanguagemodelbehaviors} \newline
\citet{wei2024simplesyntheticdatareduces} \newline
\citet{denison2024sycophancysubterfugeinvestigatingrewardtampering} \\ 
\midrule

\textbf{Direct \newline Questioning} 
& 
\textit{Use queries} like ``Are you sure about that?'' to see if the model changes correct answers to incorrect ones to please the user. 
& 
(i) Minimal prompt engineering setup;   
(ii) Simple evaluation;   
(iii) Simulates real-world user interaction. 
& 
(i) Persona context is missing;   
(ii) Binary notion of agreement might oversimplify;   
(iii) No distinction to common robustness evaluations. 
& 
\citet{sharmatowards} \newline
\citet{chen2024yesmentruthtellersaddressingsycophancy} \\ 
\midrule

\textbf{Keyword/Query Misdirection} 
&  
\textit{Deliberately insert misleading terms} into queries to test whether the model changes its response.
& 
(i) Isolate specific triggers or keywords;  
(ii) Minimal implementation;
(iii) Quantifies robustness.
& 
(i) Poor ecological validity;   
(ii) No adaptation to user persona;
(iii) No distinction to common robustness evaluations.
& 
\citet{rrv2024chaoskeywordsexposinglarge} \newline
\citet{ranaldi2024largelanguagemodelscontradict} \\
\midrule

\textbf{Visual \newline Misdirection \newline (Multimodal)} 
& 
\textit{Show an image} and a text prompt together with contradictory or misleading statements to check if the model agrees. 
& 
(i) Can test multimodal tasks;   
(ii) Ecological validity of false user input;
(iii) Evaluate model's ability to correct.
& 
(i) Requires advanced multimodal LLMs;  
(ii) Confounds with multiple phenomena;   
(iii) Resource-intensive. 
& 
\citet{lim2024measuringagreeablenessbiasmultimodal} \newline
\citet{zhao2024analyzingmitigatingsycophancylarge} \\
\midrule

\textbf{LLM-based Evaluation} 
& 
\textit{Use a (second) language model} to label sycophancy, evaluate model outputs or provide a baseline. 
& 
(i) Highly scalable;   
(ii) Continuous evaluation possible; 
(iii) simple implementation via API calls. 
& 
(i) Biases of LLM-based judge;  
(ii) Poor experiment control over LLM-based judge;  
(iii) Poor ecological validity. 
& 
\citet{williams2024multiobjectivereinforcementlearningai} \newline
\citet{chen2024openworldevaluationretrievingdiverse} 
 \\ 
\bottomrule
\end{tabular}
\caption{Comparison of common sycophancy measurement approaches in LLMs. Each of these approaches operationalizes the concept of sycophancy differently. This table subsumes five core mechanisms and summarizes their opportunities and challenges.}
\label{tab:sycophancy_strategies}
\end{table*}

\end{document}